\documentclass[10pt,twocolumn,letterpaper]{article}

\usepackage{multirow}
\usepackage[ruled,linesnumbered]{algorithm2e}
\usepackage{mathrsfs}
\usepackage{romanbar}
\usepackage{algpseudocode}

\usepackage[pagenumbers]{cvpr} 

\usepackage[dvipsnames]{xcolor}

\usepackage{multirow}

\def\bx{\mathbf{x}}
\def\bw{\mathbf{w}}
\def\bW{\mathbf{W}}
\def\bG{\mathbf{G}}

\def\bQ{\mathbf{Q}}
\def\bg{\mathbf{g}}
\def\bz{\mathbf{z}}
\def\bu{\mathbf{u}}
\def\bv{\mathbf{v}}
\def\bl{\mathbf{l}}

\DeclareMathOperator*{\argmin}{arg\,min}
\DeclareMathOperator*{\minnn}{min}

\definecolor{cvprblue}{rgb}{0.21,0.49,0.74}
\usepackage[pagebackref,breaklinks,colorlinks,citecolor=cvprblue]{hyperref}

\def\bx{\mathbf{x}}
\def\bw{\mathbf{w}}
\def\bW{\mathbf{W}}
\def\bG{\mathbf{G}}

\def\bQ{\mathbf{Q}}
\def\bg{\mathbf{g}}
\def\bz{\mathbf{z}}
\def\bu{\mathbf{u}}
\def\bv{\mathbf{v}}
\def\bl{\mathbf{l}}

\title{Hyperspherical Classification with Dynamic Label-to-Prototype Assignment}
\author{Mohammad Saeed Ebrahimi Saadabadi$^{1}$, Ali Dabouei$^{2}$, Sahar Rahimi Malakshan$^{1}$, Nasser M. Nasrabadi$^{3}$\\
$^{1,3}$ West Virginia University, $^{2}$Carnegie Mellon University\\
{\tt\small $^{1}$\{me00018, sr00033\}@mix.wvu.edu, $^{2}$adabouei@andrew.cmu.edu, $^{3}$nasser.nasrabadi@mail.wvu.edu}
}
\begin{document}
\maketitle
\begin{abstract}

Aiming to enhance the utilization of metric space by the parametric softmax classifier, recent studies suggest replacing it with a non-parametric alternative.
Although a non-parametric classifier may provide better metric space utilization, it introduces the challenge of capturing inter-class relationships. A shared characteristic among prior non-parametric classifiers is the static assignment of labels to prototypes during the training, \ie, each prototype consistently represents a class throughout the training course.
Orthogonal to previous works, we present a simple yet effective method to optimize the category assigned to each prototype (label-to-prototype assignment) during the training. To this aim, we formalize the problem as a two-step optimization objective over network parameters and label-to-prototype assignment mapping. We solve this optimization using a sequential combination of gradient descent and Bipartide matching.
We demonstrate the benefits of the proposed approach by conducting experiments on balanced and long-tail classification problems using different backbone network architectures.
In particular, our method outperforms its competitors by 1.22\% accuracy on CIFAR-100, and 2.15\% on ImageNet-200 using a metric space dimension half of the size of its competitors. 
\href{https://github.com/msed-Ebrahimi/DL2PA_CVPR24}{Code}

\end{abstract}    
\section{Introduction}
\label{sec:intro}
Image classification is a fundamental problem in deep learning \cite{rawat2017deep}. 
Deep image classification models conventionally consist of a stack of non-linear feature extractors (backbone) together with a classification layer \cite{he2016deep,le2014distributed,yang2018convolutional}. Despite diversified model design for their backbone, from convolutional neural networks \cite{he2016deep,krizhevsky2012imagenet} to transformer architecture \cite{dosovitskiy2020image,liu2021swin}, Parametric Softmax Classifier (PSC) with Cross-Entropy (CE) loss has been the \textit{de facto} choice of design  
 for the classification layer \cite{wang2022visual,he2016deep,lecun2015deep}.
This widely adopted framework enables joint training of the PSC and backbone through gradient-based learning algorithms \cite{beery2018recognition}. Yet, it suffers from several shortcomings \cite{pang2019rethinking,kasarla2022maximum,wang2022visual,yang2022inducing}.
\begin{figure}[]
\begin{center}
\includegraphics[width=1.0\linewidth]{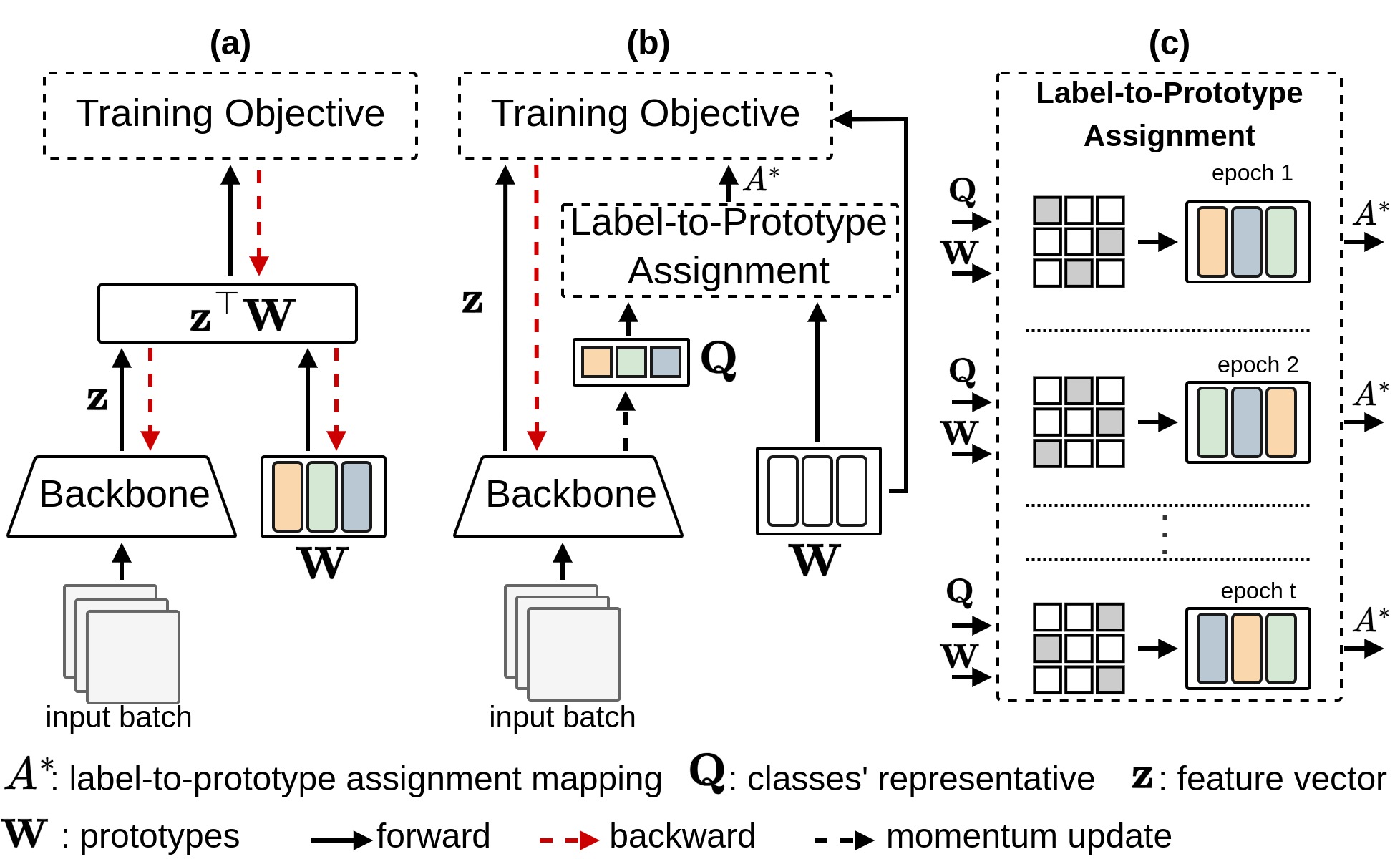}
\end{center}
\caption{
Comparison of the proposed method with the conventional PSC and the previous fixed classifier setup, using a toy example with three classes. Each color denotes a distinct class.
a) Label-to-prototype assignment remains static during training. In PSC, optimization focuses on the network, consisting of the backbone and prototypes $\bW$. In the case of a fixed classifier, only the backbone is optimized, and prototypes remain fixed.
b) In the proposed method, prototypes within the hypersphere are fixed, and optimization targets the backbone and the label that each prototype represents.
c) Toy example showing changes in label-to-prototype assignment during training.
}\label{fig1}
\vspace{-3mm}
\end{figure}

{First}, it overlooks the intra-class compactness, focusing solely on the optimization of the relative intra and inter-class distances \cite{pang2019rethinking,wang2021exploring}.
{Second}, it fails to fully exploit the metric space, leading to a localized solution \cite{kasarla2022maximum,liu2018decoupled,duan2019uniformface}.
{Third}, the standard PSC enforces a fixed dimensionality of the output space, which hampers transferability and results in linear growth of the classifier parameter associated with the number of classes \cite{shen2023equiangular,wang2022visual}. {Forth}, when facing an unbalanced dataset \textit{w.r.t.} sample per class, the PSC struggles with suboptimal performance for minority classes \cite{fang2021exploring,deng2021variational,yang2022inducing}.
To address these issues, studies suggest employing auxiliary supervisions \cite{duan2019uniformface,wen2016discriminative}, sample-based objectives \cite{schroff2015facenet,deng2021variational}, distributed training \cite{an2021partial,an2022killing}, active class selection \cite{goodman2001classes}, and loss reweighting \cite{cui2019class,khan2019striking}. 
However, these methods concentrate exclusively on addressing a single limitation and do not generalize to all the issues \cite{wang2022visual}.

Recently, studies have focused on replacing PSC with non-parametric solutions \cite{wang2022visual,yang2022inducing, mettes2019hyperspherical}.
Yang \etal \cite{yang2022inducing} employed a fixed Equiangular Tight Frame (ETF) geometrical structure as prototypes of the PSC providing maximal inter-class separation, optimal metric space exploitation, and a constant number of trainable parameters as the number of classes increases. Furthermore, fixed prototypes alleviate the problem of `\textit{minority collapse}' in the learnable classifier trained on an imbalanced dataset \cite{fang2021exploring}. 
However, a major issue with the ETF classifier \cite{yang2022inducing} is that ETF prototypes force any pairs of classes to reflect equal similarity.
Intuitively, semantically unrelated classes should be farther apart from the related ones, and equal inter-class similarity for all categories is counter-intuitive \cite{khosla2020supervised}. Besides, ETF requires the metric space dimensionality to be greater than the number of classes, limiting its applicability \cite{papyan2020prevalence,lu2020neural}.

In the quest for a more appropriate non-parametric classification alternative, Wang \etal \cite{wang2022visual} obtain fixed prototypes by summarizing each class into a cluster. Albeit effective
for supervising intra-class compactness, it fails to fully exploit the metric space. 
Recently, Mettes \textit{et al.} \cite{mettes2019hyperspherical} utilized fixed equidistributed hyperspherical points as prototypes of PSC.
While equidistributed prototypes enhance metric space exploitation \cite{kasarla2022maximum,duan2019uniformface,mettes2019hyperspherical}, the question that naturally arises here is: \textit{with fixed prototypes, how to define the label-to-prototype assignment to capture the dataset's inter-class relationships?}
To answer this, HPN \cite{mettes2019hyperspherical} utilizes the classes' names as privileged information; however, such privileged information can be inadequate or absent \cite{ghadimi2021hyperbolic}. A common characteristic amongst previous learnable and fixed classifiers is that the label-to-prototype assignment is static, and each prototype represents a unique label throughout the training course, as depicted in Figure \ref{fig1}a.

In this study, our methodology diverges from static label-to-prototype assignment and focuses on optimizing the assignment throughout the training when the prototypes are fixed, as illustrated in Figure \ref{fig1}b.
Therefore, the model changes the label-to-prototype assignment while the prototypes are pre-distributed in the hypersphere and fixed. Ideally, this ensures the maximal inter-class separation throughout fully exploiting the metric space in a manner consistent with the inter-class relationships, as depicted in Figure \ref{fig1}c. 
We divide the $\small{c}$-way classification into three sub-tasks: 1) organizing $\small{c}$ equidistributed prototypes in the hypersphere, 2) solving a bipartite matching problem from the classes to the predefined prototypes, and 3) solving a regression problem from input to discrete approximation of unit hypersphere. 
The solution to the former depends merely on the metric space dimensionality and the number of classes.
We formulate the second and final stages as a two-step optimization over the label-to-prototype assignment mapping and network parameters.  
The main contributions of this paper can be summarized as follows:

\begin{itemize}
    \item We propose a novel classification framework in which the prototypes are fixed but the label-to-prototype assignment is adaptive and changes during the course of training.  
    
    \item We formalize the problem as a two-step optimization and solve it using a combination of bipartite matching and gradient descent. This procedure does not use any auxiliary information to achieve this goal. 

    \item We empirically show that our method is a generalized version of the ETF methods by removing the constraint on the number of classes. 
    
    \item We evaluate our method on both balanced and long-tailed classification, achieving superior or comparable results compared to state-of-the-art fixed classifier methods while being more effective in training and scale. 
\end{itemize}

\section{Related Works}

\label{sec:relatedworks}
Deep Neural Networks (DNNs) have become widely used in machine learning, excelling in classification with end-to-end training using PSC \cite{he2016deep,le2014distributed}.
Motivated by the huge computational demand of the last linear layer \cite{hoffer2018fix,shen2023equiangular,wang2022visual}, limited inter-class separation \cite{duan2019uniformface,mettes2019hyperspherical,pang2019rethinking,wang2021exploring}, and poor classification accuracy of minority classes \cite{deng2021variational,yang2022inducing,shen2023equiangular,wang2022visual}, recent studies have focused on replacing the PSC with non-parametric alternatives \cite{hoffer2018fix,mensink2013distance,movshovitz2017no,guerriero2018deepncm,li2019prototype}.
 To this end, a group of studies \cite{lu2020neural,yang2022inducing,yang2023neural} focused on replacing prototypes of PSC with predefined fixed ETF. A set of $\small{c}$ lines $\small{\mathcal{L}=\{\bl_1,\bl_2,...,\bl_c\}}$ through the origin of $\small{d}$-dimensional space is termed equiangular if there is a constant $\small{\rho \geq 0}$, such that $\small{\left< \bl_i,\bl_j\right>= \pm \rho}$ for all $\small{1\leq i<j\leq c}$ \cite{zhu2021geometric}. Here, $\small{\left< ., . \right>}$ represents the inner product.
While ETF provides optimal inter-class separation, we argue that it can be sub-optimal from the perspective of inter-class relationships.
Intuitively, unrelated classes should be farther apart from the related ones.  
However, an ETF classifier forces a separable but counter-intuitive structure in which all the classes have equal relationships.
Moreover, while constructing simplex ETF for pair $\small{(d,c)}$ is not limited to $\small{d\geq (c-1)}$, studies demonstrate that ETF is only applicable to DNNs classification framework when $\small{d\geq (c-1)}$, which critically limits its practical utilization \cite{sustik2007existence,balla2018equiangular,zhu2021geometric}.

To bypass these limitations, Wang \etal \cite{wang2022visual} adhere to the well-established idea of the nearest centroid classifier. The classifier in this framework is obtained by summarizing each class into a cluster. While effective in capturing inter-class relationships, the efficacy of metric space exploitation is overlooked \cite{mettes2019hyperspherical}.
Mettes \etal \cite{mettes2019hyperspherical} proposed employing fixed equidistributed hyperspherical points as prototypes of PSC. Ideally, equidistributed prototypes provide full exploitation of the metric space \cite{ duan2019uniformface, mettes2019hyperspherical}.
While promising, this method requires privileged information, to capture the inter-class relationships, which can be inadequate or absent \cite{ghadimi2021hyperbolic}. Common among previous methods is that the label-to-prototype assignment is constant during training.

In an orthogonal direction, we propose to fix the prototype in an equidistributed organization and change the label-to-prototype assignment. Thus, the semantic category that every prototype represents is not unique during the training and changes \textit{w.r.t.} the state of the feature extractor. For distributing the prototypes on the hypersphere, we optimized a cost function based on the Gaussian potential kernel, which is closely connected to universally optimal point configurations \cite{borodachov2019discrete,cohn2007universally}, and we empirically show that our equidistributed answer converges to the ETF structure when the metric space dimensionality requirement is met.
\section{Method}
\label{sec:method}

\textbf{Notation.} Let $\small{\mathcal{D}=\left\{ (\bx_i, y_i) \in \mathcal{X} \times \mathcal{Y} \right\}_{i=1}^{n}}$ be the training set consisting of $\small{c}$ classes and $\small{n}$ samples. Suppose $M_{\theta}(.):\mathcal{X}\rightarrow\mathbb{R}^d$ is the deep feature extractor that maps an input sample $\bx_i$ to $\small{d}$-dimensional representation $\bz_i \in \mathbb{R}^d$. $\small{{\bW}=[\bw_{{1}},\bw_{2}, \dots,\bw_{c}] \in  \mathbb{R}^{d\times c}}$ is the matrix storing hyperspherical prototypes $\bw_i \in \mathbb{R}^{d}$. $A: [1,2, \dots, c] \rightarrow [1,2, \dots, c]$ is the label-to-prototype assignment, denoting a bijective mapping from labels to prototypes. $A$ is constant in conventional PSC and equiangular setup and optimizable in our method.
$\small{{\bQ}=[\bar{\bz}_1,\bar{\bz}_2, \dots, \bar{\bz}_c]\in \mathbb{R}^{d\times c}}$ is the matrix of the classes' representative, storing the momentum-averaged latent feature of each class during training.
For the convenience of presentation, all the representations and prototypes are $\ell_2$-normalized.

\begin{figure}[]
\begin{center}
\includegraphics[width=0.9\linewidth]{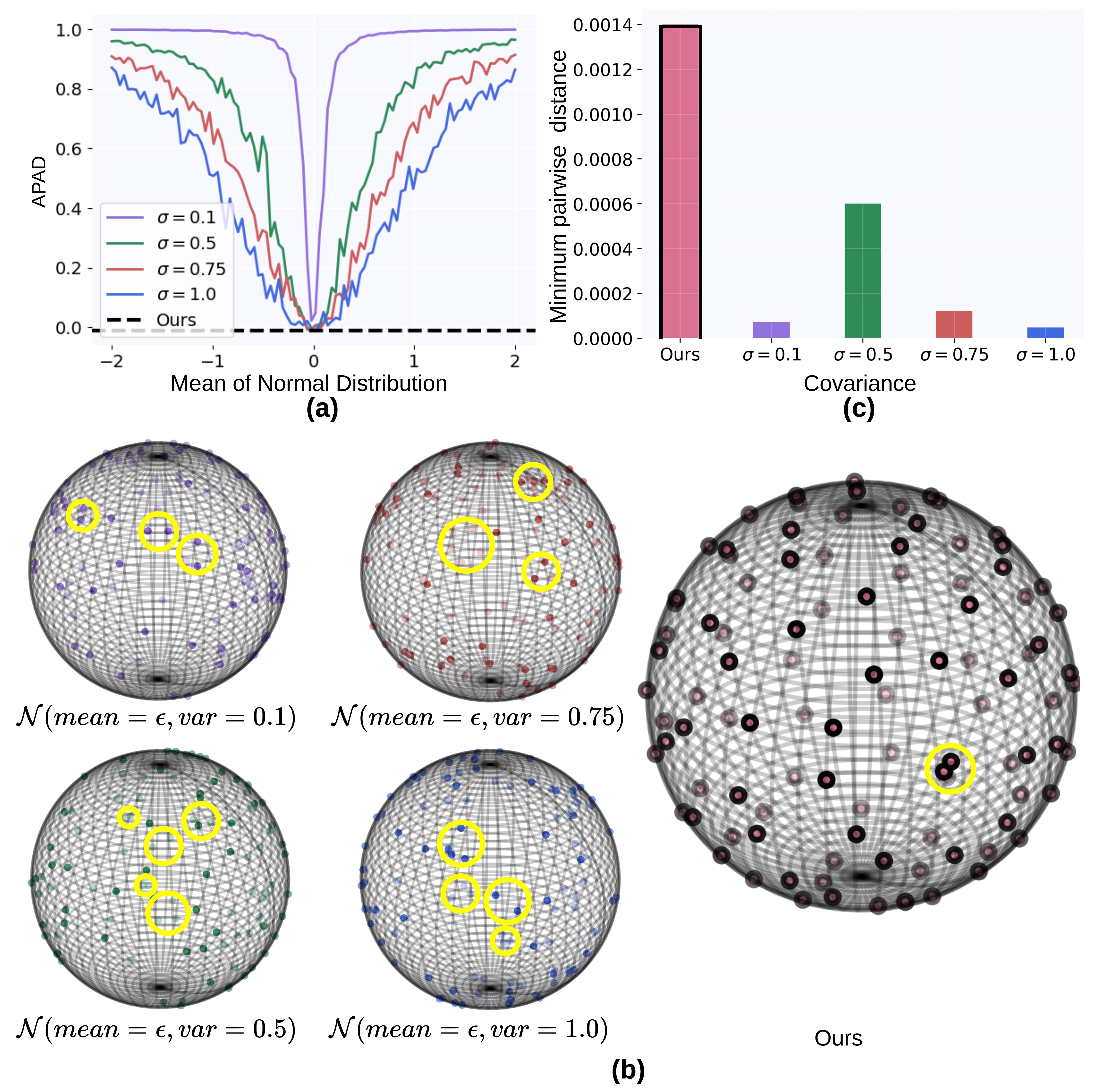}
\end{center}
\vspace{-16pt}
\caption{\small{
Comparing the Average Pairwise Angular Distance (APAD) value of the 100 prototypes drawn from multivariate Gaussian distributions with the covariance matrix of $\sigma I$ and the proposed optimized prototypes. 
a) Analysis of the APAD value. Notably, random prototypes drawn from arbitrary zero-mean distributions yield the optimal APAD value, underscoring that uniformity across $\small{S^{d-1}}$ cannot be solely guaranteed by this objective function. b) Illustrating optimized and degenerate solutions on the $\small{S^2}$. Highlighted in yellow are areas where multiple prototypes exhibit closer proximity than ideal. c) Comparison of the minimum cosine distance, \ie, $1-\cos(\bw_i,\bw_j)$, for our optimized and degenerate solutions prototypes. Greater distances are indicative of superior utilization of metric space. }}\label{APAD10}
\vspace{-16pt}
\end{figure}

\subsection{Overview}
We are interested in learning a classification model that fully exploits the metric space while capturing the underlying characteristics of the target dataset. Inspired by the recent works on replacing the PSC with non-parametric alternatives \cite{hoffer2018fix,mensink2013distance,movshovitz2017no,guerriero2018deepncm,jetley2015prototypical,li2019prototype},
 we fix the equidistributed hyperspherical prototypes  \cite{yang2023neural,shen2023equiangular} to maintain large-margin separation among class prototypes during the training. Hence, the supervised classification is generally defined as mapping input images to a discrete approximation of the unit hypersphere:
\begin{equation}\label{bilevel1}
 \small
 \begin{aligned}
  \mathop{{\minnn}}_{{\theta}} \sum_{i=1}^{n}L(M_{\theta}(\bx_i), \bw_{A(y_i)}),
\end{aligned}
\end{equation}
where $\small{L}$ is the training objective, tryings to align features $\bz_i$ with their corresponding prototype $\bw_{A(y_i)}$. 

\begin{figure*}[t]
\centering
\includegraphics[width=1.0\linewidth]{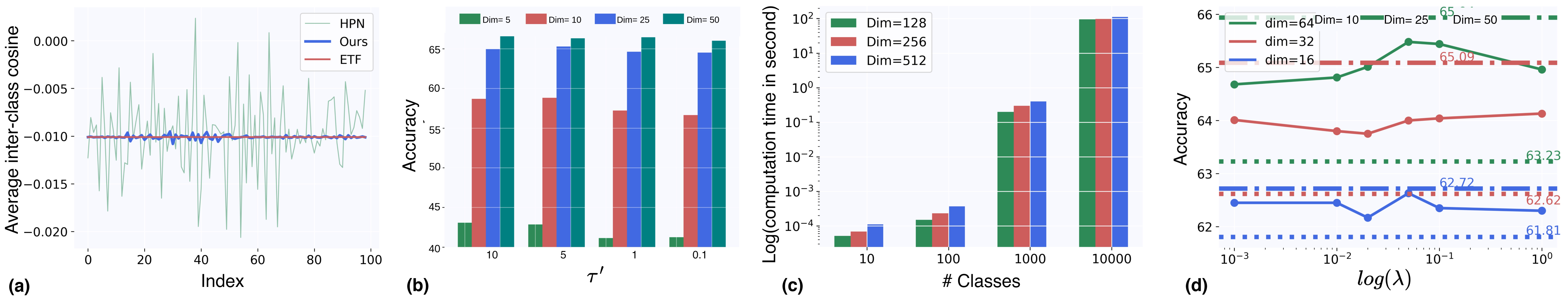}
\vspace{-16pt}
\caption{\small{a) Average inter-prototype cosine when $\small{d=c=100}$.  b) Classification accuracy (\%) on CIFAR-100 with ResNet-32 when the $\tau'$ changes. c) Time consumed for updating the $A$. Since we update the label-to-prototype assignment every epoch,  $\tau'=1.0$, this computation time is negligible compared to the total training time. d) Effect of regularizing the PSC with $\small{L_{uni}}$ with different scaling hyperparameter $\small{\lambda}$. The horizontal dotted and '-.-' lines represent the PSC and proposed method performance, respectively.
}}\label{fig3}
\vspace{-16pt}
\end{figure*}

In the case of ETF \cite{yang2022inducing}, the assignment of prototypes is an arbitrary mapping that remains constant during training, as any given pair of prototypes exhibits equal similarity. 
While this assumption may hold for specific data, it fails to generalize, \eg, for a multi-modal data distribution where different classes share similar properties. Thus, it is expected that the similarity between members of the same super-class to be higher than the similarity between members of different super-classes \cite{bojanowski2017unsupervised}, \eg, as the `\textit{truck}' and `\textit{automobile}' classes in CIFAR-10.
Also, in the case of PSC setup, $A$ is the identity mapping $A(y_i)=y_i$ and fixed during the training, \ie, each class is assigned to a specific trainable prototype. However, PSC ignores the distribution of features in holistic metric space and may lead to high locality \cite{duan2019uniformface, mettes2019hyperspherical}. Here, $A$ is not fixed during the training, and we aim to find the best prototype for each class from the predefined prototype set $\bW$. Hence, our main objective is: 
\begin{equation}\label{bilevel3}
 \small
 \begin{aligned}
  \mathop{{\minnn}}_{{\theta}} \mathop{{\minnn}}_{{A}} \sum_{i=1}^{n}L(M_{\theta}(\bx_i), \bw_{A(y_i)}).
\end{aligned}
\end{equation}

Our overall framework involves an optimization problem \textit{w.r.t.} two sets of variables $\small{\theta}$, and $A$. 
This problem does not have a closed-form solution due to the highly non-linear nature of DNNs. Furthermore, $A$ is not differentiable, and the whole framework cannot be optimized using gradient descent. We formulate a two-step optimization to solve this problem. In the first step, we solve a bipartite matching problem between classes' representatives $\bQ$ and fixed prototypes $\bW$, which is essential to capture the inter-class relationships in the metric space.
In second step, we optimize $\small{L}$ with respect to $\small{\theta}$ given $\small{{A}}$ and $\small{{\bW}}$.

Inspired by \cite{wang2020understanding}, we enhance the metric space utilization by discretizing the hypersphere with prototypes distributed as uniformly as possible prior to training.
During the training, optimization in Equation \ref{bilevel3} does not involve updating the prototypes and maintains the large-margin separation among classes.  
Albeit the predefined and fixed position of $\small{{\bW}}$, our framework changes the corresponding label for each prototype to reduce the classification loss function. 
In this way, given well-separated prototypes, our approach assigns classes to prototypes to minimize the classification loss and capture inter-class dependencies rather than fixing class relationships similar to the ETF setup
\cite{yang2022inducing}. Furthermore, the label-to-prototype assignment is based on model learning with no need for privileged information \cite{mettes2019hyperspherical}. Algorithm \ref{alg1} provides an overview of the presented method.

\subsection{{Classification with Dynamic Label-to-Prototype Assignment}}\label{mainmethod}
Here, we provide a detailed description of our method. Prior to training, we distribute hyperspherical prototypes as uniformly as possible in a dataset and architecture-independent manner. Given the optimized prototypes, we adopt a two-step algorithm where the goal is to find parameters $\theta$ and $A$ that best describe the data distribution.

\subsubsection{Hyperspherical Prototype Estimation}
Metric space utilization of the PSC relies on its hyperspherical prototype distribution \cite{duan2019uniformface}. To enhance metric space utilization, studies tried to encourage the uniformity of prototypes' distribution in the hyperspherical space by adding regularization to conventional PSC setup \cite{duan2019uniformface,liu2018learning,wen2016discriminative}. Recently, Yang \etal \cite{yang2022inducing} defined prototypes as vertices of a simplex ETF that provides maximal inter-class separability. However, the ETF structure of prototypes is a strong assumption in which all pairs of classes are forced to have equal similarity. Also, ETF requires $d\geq (c-1)$, impractical for large $c$ \cite{shen2023equiangular,yang2022inducing}. 
To maximize the metric space utilization, we distribute prototypes in the hypersphere as uniformly as possible in a data and architecture-independent manner and fix them during the network training.

For $\small{d=2}$ and $\small{c}$ prototypes, the problem of uniformly distributing prototypes on the unit hypersphere $\small{{S}^{d-1}}$ reduces to splitting the unit circle into equal slices with \(\dfrac{2\pi}{c}\) angles. 
However, no optimal solution exists for \(d\geq 3\) \cite{saff1997distributing}. As a viable alternative, we use a differentiable uniformity metric to encourage mapping prototypes to a uniform distribution over $\small{{S}^{d-1}}$. Designing an objective minimized by uniform distribution is not trivial \cite{cohn2007universally}. For instance, Wang \etal \cite{wang2020understanding} demonstrated that an arbitrary zero mean distribution is a degenerate solution of Average Pairwise Angular Distance (APAD) \cite{borodachov2019discrete,cohn2007universally}, as depicted in Figure \ref{APAD10}.

\begin{table*}[]
\addtolength{\tabcolsep}{6pt} 
\small

\begin{center}
\begin{tabular}{lcccc|cccc}
\toprule

\multicolumn{1}{l}{\multirow{2}{*}{}} & \multicolumn{4}{c|}{CIFAR-100}                                                                                                                    & \multicolumn{4}{c}{ImageNet-200}                                                                                                                         \\  

                        & $\small{d=10}$                                & $\small{d=25}$                                & $\small{d=50}$                                & $\small{d=100}$                               & $\small{d=25}$                                & $\small{d=50}$                                & $\small{d=100}$                                  & $\small{d=200}$                               \\ \midrule
PSC                                     & 25.67                                   & 60.0                                   & 60.6                                   & 62.1                                & -                                   & -                                   & -                                      & 33.1                                \\
Word2Vec \cite{mikolov2013distributed}                                   & 29.0                                & 44.5                                & 54.3                                & 57.6                                & 20.7                                & 27.6                                & 29.8                                   & 30.0                                \\
HPN \cite{mettes2019hyperspherical}                                        & 51.1                                & 63.0                                & 64.7                                & \textbf{65.0}                                & 38.6                                & 44.7                                & 44.6                                   & \textbf{44.7}                                \\ 
Ours                                        & \textbf{57.21} & \textbf{64.63} & \underline{\textbf{66.22}}  & {62.85}  & \textbf{41.71}  & \textbf{46.57}  & \underline{\textbf{46.85}}& 37.28  \\  \bottomrule
\end{tabular}
\end{center}
\vspace{-16pt}
\caption{\small{Classification accuracy (\%) of our method and baseline prototypical methods using ResNet-32. Please note that our best result across dimensions (underlined), is always better than the best performance of baseline methods.
Our method obtains the best accuracy across dimensions and datasets when $\small{d< c}$, emphasizing the role of dynamic label-to-prototype mapping in metric space exploitation.}} 
\label{table1}
\vspace{-8pt}
\end{table*}

 \begin{table}[]
\addtolength{\tabcolsep}{-4pt} 
\small

\begin{center}
 \resizebox{0.9\linewidth}{!}{
\begin{tabular}{lccccc}
\toprule
Dataset      & PSC  & Word2Vec \cite{mikolov2013distributed} & HPN \cite{mettes2019hyperspherical}  & ETF \cite{yang2022inducing}  & Ours  \\ \midrule
CIFAR-100   & 62.1 & 57.6     & 65.0 & 65.34 & \textbf{66.22} \\
ImageNet-200& 33.1 & 30.0     & 44.7 & 37.15     & \textbf{46.85} \\ \bottomrule
\end{tabular}}
\end{center}
\vspace{-10pt}
\caption{\small{Comparison of classification accuracy (\%) of the proposed method with $d=\frac{c}{2}$ and other prototypical methods with $d=c$ using ResNet-32.}} \label{cifar100etf}
\vspace{-10pt}
\end{table}

We built our uniformity objective upon recent progress in contrastive representation learning \cite{wang2020understanding}. Let $\small{p_{{\bw}}(.)}$ be the prototypes' distribution over $\small{{S}^{d-1}}$. $G_t(\bu,\bv)\triangleq S^{d-1} \times S^{d-1} \rightarrow \mathbb{R}_{+}$ is the Gaussian potential kernel: 
\begin{equation}\label{GPK}
 \small
 \begin{aligned}
G_t(\bu,\bv)\triangleq e^{-t||\bu-\bv||_2^2}; \quad t>0,
\end{aligned}
\end{equation}
the uniformity loss can be defined as:
\begin{equation}\label{ALGPK}
 \small
 \begin{aligned}
	L \triangleq \log{\mathop{\mathbb{E}}_{\bw_i,\bw_j \mathop{\backsim}\limits^{\text{i.i.d.}} p_{\bw}}}[G_t(\bu,\bv)];\quad t>0,
\end{aligned}
\end{equation}
 which is nicely tied with the uniform distribution of points on the $\small{S^{d-1}}$ \cite{bochner1933monotone}; detailed derivation can be found in \cite{wang2020understanding}.

In practice, we first randomly initialize the prototype matrix $\small{{\bW}\in \mathbb{R}^{d\times c}}$. At every iteration, we optimize a subset of $\small{{\bW}}$ using the following loss function:
\begin{equation}\label{practical_loss}
 \small
 \begin{aligned}
	L_{uni} = \log{(\frac{1}{\hat{c}} \sum_{i=1}^{\hat{c}}\sum_{j=1}^c {g_{i,j}})},
\end{aligned}
\end{equation}
where $\small{\widehat{{\bW}} \in \mathbb{R}^{d \times \hat{c}}}$ is a subset of $\small{{\bW}}$ that randomly selects $\small{\hat{c} < c}$ prototypes, and  $\small{g_{i,j}}$ is the element of $\small{\bG}$ that reflects the pairwise Gaussian potential between prototypes $\small{\bw_i}$ and $\small{\bw_j}$. Subsequent to updating the prototypes, they are mapped to the unit hypersphere using $\ell_2$-normalization.

Please note that the solution to the distribution of the prototypes is independent of the dataset and network architecture. Instead, it depends merely on the metric space dimensionality, $d$, and the number of classes, $c$. Once we obtain a solution for a specified $d$ and $c$, this ${\bW}$ can be used across problems with the same number of categories and the same metric space dimensionality.
For instance, in Table \ref{table2} the same ${\bW}$ is used for CIFAR-10, SVHN, and STL-10.

\subsubsection{Input-to-Prototype Mapping}\label{lipm}
As pointed out by \cite{deng2021variational, yang2022inducing}, in the training of PSC, the `\textit{pull-push}' mechanism of Cross-Entropy (CE) implicitly forces intra-class compactness and inter-class separability, please see Section \ref{fl} of Supplementary Material for detailed analysis on CE derivative. Our method fixes the prototypes as the solution for inter-class separability, and the model does not learn it during the training. Hence, for a training sample $\bx_i$ we seek to learn a mapping from the input to its assigned prototype, \ie, `\textit{pull}' $\bz_i$ toward $\bw_{A(y_i)}$:
\begin{equation}\label{cosineSimilarity}
 \small
 \begin{aligned}
 L_{IPM}(\bz_i, \bw_{A(y_i)}) = \frac{1}{2} {(\bz_i^\top\bw_{A(y_i)}-1)^2},
\end{aligned}
\end{equation}
this loss is only concerned with solving a regression task between the $\small{\bz_i}$ and $\small{\bw_{A(y_i)}}$; supervising the intra-class compactness by decreasing the angular distance of samples and their assigned prototypes. 

Unlike the CE loss, Equation \ref{cosineSimilarity} is derived using a single prototype per sample. This approach reduces the computational workload compared to CE, where all prototypes are required for every sample \cite{an2021partial}.
Since we do not update the prototype during the training, our approach requires a back-propagation step through the $M_\theta$. The partial derivative of Equation \ref{cosineSimilarity} can be given as:
\begin{equation}\label{gardCosSim}
 \small
 \begin{aligned}
 \frac{\partial L_{IPM}}{\partial \bz_i} = - (1-\bz_i^\top\bw_{A(y_i)})\bz_i^\top\bw_{A(y_i)},
\end{aligned}
\end{equation}
which is identical to the `\textit{pull}' force in the CE derivative.
Since the Equation \ref{gardCosSim} is only concerned with aligning the samples to their assigned prototype, the problem of frequent \textit{passive update} \cite{deng2021variational} of minority classes is relieved for long-tail datasets \cite{fang2021exploring}.
Ideally, the combination of Equation \ref{cosineSimilarity} and fixed equidistributed prototypes, increases the intra-class compactness in an optimally separate prototype organization, \ie, maximal inter-class separation.

\subsubsection{Bijective Label-to-Prototype Mapping}
Our proposal hinges on finding suitable label-to-prototype assignment $A$ during the training.
In the ETF setup, $A$ is initialized randomly. In the case of PSC setup, the assignment is identity mapping. Note that in both these scenarios, $A$ remains constant during the training. 
Here, our objective is to find assignment mapping that maximizes the log-likelihood function of the observed class:
\begin{equation}\label{bilevel5}
 \small
 \begin{aligned}
   A^* = & \argmin_{A}\sum_{j=1}^{c}{-\log{\frac{e^{\bar{\bz}_j^{\top} \bw_{A(j)}}}{\sum_{k=1}^{c}{e^{\bar{\bz}_j^{\top} \bw_{A(k)}}}}}}\\
       = & \argmin_{A}\sum_{j=1}^{c}\left[{- \bar{\bz}_j^{\top} \bw_{A(j)} + \log{{\sum_{k=1}^{c}{e^{\bar{\bz}_j^{\top} \bw_{A(k)}}}}}}\right],
\end{aligned}
\end{equation}
where $\bar{\bz}_{y_i} = \alpha \bar{\bz}_{y_i} + (1-\alpha) \bz_i$ is the momentum averaging over the representations of each class and $\small{\alpha} \in [0,1]$ is the momentum coefficient  \cite{he2020momentum,lucas1990exponentially}. The term $\log{\sum_{k=1}^{c}e^{\bar{\bz}_j^{\top} \bw_{A(k)}}}$ is constant for all elements of $A$. Hence, the optimization term becomes: 
\begin{equation}\label{bilevel6}
 \small
 \begin{aligned}
   A^* = \argmin_{A}\sum_{j=1}^{c}{-{{{\bar{\bz}_j^{\top} \bw_{A(j)}}}}},
\end{aligned}
\end{equation}
following \cite{hu2021vivo,liu2021end} we solve this bipartite matching problem using Hungarian algorithm \cite{kuhn1955hungarian}.
The resulting $A^*$ is used to optimize Equation \ref{cosineSimilarity}. We update the prototypes' assignment every $\tau$ iterations to sync it with model learning.

\section{Experiment}
\label{sec:experiment}

\begin{algorithm}[t]
\footnotesize	
\caption{Classification with dynamic Label-to-prototype assignment}\label{alg1}
Initialize ${A}$, $\small{{\bW}}$, $\small{\theta}$,  $\small{{\bQ}}$, $\small{t_1>0}$, and $\small{t_2>0}$.

\For{$\small{t=0}$ ... $\small{t_1\!-\!1 }$}{  

Compute $\small{L_{uni}}$, $\small{\forall \bw_i \in {\bW}}$ 

Update $\small{{\bW}}$ using $\small{\nabla_{{\bW}}L_{uni}}$\Comment{\textcolor{blue}{prototype estimation}}

}

\For{$\small{t=0}$ ... $\small{t_2\!-\!1 }$}{
\For{Batch in Dataset}{
 $\small{\bz={M}(Batch)}$
 
Update $\bQ$ using momentum averaging the representations of each class

Compute $\small{L_{IPM}}$ using $\bz$, $A$, and $\bW$

Update $\small{{M_{\theta}}}$ using $\small{\nabla_{\theta} L_{IPM}}$; \Comment{\textcolor{blue}{backbone training}}
}
Update assignment $\small{A}$ using Equation \ref{bilevel6}; 
}

\end{algorithm}

As discussed in Section \ref{lipm}, employing fixed prototypes and $L_{IPM}$ is particularly effective for the long-tail classification problem, also shown in
\cite{fang2021exploring,deng2021variational}.
Hence, we conduct experiments in balanced and long-tailed setups to illustrate the efficacy of the proposed method in both scenarios.
\subsection{Implementation Details}
For precomputing the prototypes, we used SGD optimizer with a constant learning rate of 0.1, and the optimization was performed for 1000 iterations with the batch size equal to the number of classes.
We conduct evaluations on balanced ImageNet-1K, ImageNet-200, and CIFAR-100 using ResNet-32, ResNet-50 \cite{he2016deep}, and Swin-Transformer \cite{liu2021swin} architectures. We train ResNet-32 and ResNet-50 using the SGD optimizer, and Swin-T is trained from scratch using Adam-W. For the sake of a fair comparison in balanced evaluations, we followed \cite{mettes2019hyperspherical} for data augmentation. We conduct long-tailed evaluations on CIFAR-10, CIFAR-100, SVHN, and STL-10 using ResNet-32 and on ImageNet-LT \cite{liu2019large} using ResNet-50. The models are trained from scratch using the SGD optimizer. 
The detailed experimental setup is described in Section \ref{trainingsetup} of Supplementary Material.  

\subsection{Balanced Classification}
We compare our method with three different baselines of PSC, Word2Vec \cite{mikolov2013distributed} and HPN \cite{mettes2019hyperspherical}.
Table \ref{table1} presents results for evaluations on ImageNet-200 and CIFAR-100. 
For both datasets, our method obtains the highest accuracy across metric space dimensions and datasets. This shows the better metric space exploitation by our predefined prototypes which are distributed based on Gaussian potential. Furthermore, these results demonstrate the superiority of our method in optimizing label-to-prototype assignment based on model training. Note that HPN utilizes privileged information of word embedding of classes' names which is not related to the training and comes from an independent model. However, we optimize the assignment based on model learning without employing any auxiliary information. We also observe that the superiority of the proposed method is more remarkable when the metric space dimension is significantly lower than the number of classes $d\ll c$. It can be attributed to the dynamic label-to-prototype assignment that captures inter-class relationships which are more noticeable when $d\ll c$. 

 \begin{table}[]
\small

\begin{center}
\resizebox{1\linewidth}{!}{
\begin{tabular}{lcccc}
\toprule
Method   & Venue & Backbone  & Optimizer &Accuracy (\%) \\ \midrule 
PSC \cite{he2016deep}& CVPR2016 &ResNet-50 &  SGD         & 76.51          \\
DNC \cite{wang2022visual} & ICLR2022 &ResNet-50 &  SGD         & 76.49          \\
Goto \etal \cite{goto2024learning} & WACV2024 &ResNet-50 &  SGD         & 77.19          \\
Kasarla \etal \cite{kasarla2022maximum} & NEURIPS2022 &ResNet-50 &  SGD         & 74.80          \\
 Ours    & CVPR2024 & ResNet-50 &SGD      & \textbf{77.47}          \\ \midrule 

 DNC \cite{wang2022visual} & ICLR2022 &ResNet-101 &  SGD         & 77.80          \\
 Goto \etal \cite{goto2024learning} & WACV2024 &ResNet-101 &  SGD         & 78.27         \\
 Kasarla \etal \cite{kasarla2022maximum} & NEURIPS2022 &ResNet-152 &  SGD         & 78.5          \\
 Ours    & CVPR2024 & ResNet-101 &SGD      & \textbf{79.63}          \\ \midrule
PSC & CVPR2016 &Swin-T     &  AdamW         &   76.91            \\
Ours &  CVPR2024   &Swin-T     &      AdamW             & \textbf{77.26}               \\ \bottomrule
\end{tabular}}
\end{center}
\vspace{-16pt}
\caption{\small{Classification accuracy (\%) on ImageNet-1K. The training was performed for 100 epochs when $d=512$.}} \label{imagenetresult}
\vspace{-16pt}
\end{table}

Moreover, we observe that our method achieves the best performance across metric space dimensions. Specifically, in CIFAR-100, our approach attains an accuracy of 66.22\% in $\mathbb{R}^{50}$, surpassing HPN's best performance by 1.22\%, regardless of dimension. Similarly, in ImageNet-200, our method in $\mathbb{R}^{100}$ outperforms HPN's best performance across metric space dimensions by 2.14\%.
This emphasizes the role of label-to-prototype assignment which is an optimization variable in our method and constant in other fixed classifier baselines.
When $d=c$ our optimized prototypes converge toward ETF which leads to the counter-intuitive solution in that all classes have equal similarity, causing our best performance to be achieved when $d<c$.

In terms of robustness to changes in metric space dimension, HPN experiences a performance drop of 11.89\% and 1.7\% on CIFAR-100 when transitioning from $25\rightarrow 10$ and $50\rightarrow 25$, respectively. In comparison, our method demonstrates greater stability with performance degradations of 7.41\% and 1.6\%.
Similarly, in ImageNet-200, when transitioning from $50\rightarrow 25$, HPN experiences a 6.1\% performance drop, while our method is more robust with a performance degradation of only 4.85\%. This performance robustness demonstrates that despite the reduction in metric space dimension and inter-class separability, our label-to-prototype assignment successfully captures inter-class relationships, leading to the capability to handle classification with much fewer metric space dimensions, \ie, better metric space exploitation.

Table \ref{cifar100etf} compares the results of equidistributed prototypes, \ie, Word2Vec \cite{mikolov2013distributed}, HPN \cite{mettes2019hyperspherical}, and our approach, with ETF \cite{yang2022inducing} on CIFAR-100 and ImageNet-200 evaluations. In these evaluations, other methods use $d=c$ (the best performance of baseline methods), while our method employs $d=\frac{c}{2}$. Please note that the ETF cannot implemented in $\mathbb{R}^{\frac{c}{2}}$ since the constraint for ETF requires $d\geq(c-1)$. We observe that the presented method outperforms ETF structure by 0.87\% and 9.15\% in CIFAR-100 and ImageNet-200, respectively. The results highlight the effectiveness of our approach in lifting the dimensionality constraint of the ETF and outperforming baselines with less metric space dimension.
Our results on ImageNet-1K are presented in Table \ref{imagenetresult}, showing the superior performance of the presented method when compared to the conventional PSC framework. These improvements signify that our approach is not restricted to small-scale classification like CIFAR-100 or architecture like ResNet-32.

\begin{table}[]
\addtolength{\tabcolsep}{7.0pt} 
\small
\begin{center}
\begin{tabular}{lcccc}
\toprule
Method & \multicolumn{1}{c}{$d$}   & \multicolumn{1}{c}{0.005} & \multicolumn{1}{c}{0.01} & 0.02 \\ \hline
PSC    & \multicolumn{1}{c}{128} & \multicolumn{1}{c}{38.7}  & \multicolumn{1}{c}{43.0} & 48.1 \\
ETF    & \multicolumn{1}{c}{128} & \multicolumn{1}{c}{\underline{40.9}}  & \multicolumn{1}{c}{\textbf{45.3}} & 50.4  \\ \hline
Ours   & 128                     & \textbf{41.3}                      & 44.9                     & \underline{50.7}  \\
Ours   & 50                      & \underline{40.9}                      & \underline{45.2}                     & \textbf{50.8}  \\ \bottomrule
\end{tabular}
\end{center}
\vspace{-16pt}
\caption{\small{ Long-tailed CIFAR-100 classification accuracy (\%) of the proposed and baseline prototypical methods using ResNet-32. Note that our method using $d=50$ could get the best or second-best performance in all the experiments.
}} \label{stable14}
\vspace{-7pt}
\end{table}

\subsection{Long-tail Classification }

The standard PSC framework with Cross-Entropy (CE) loss suffers from frequent `\textit{passive updates}' of minority classes, leading to low separability among them \cite{deng2021variational}. Table \ref{stable14} presents the results for long-tail evaluations on CIFAR-100. Our method consistently outperforms the standard PSC classifier across various imbalance factors, demonstrating its ability to alleviate the issue of frequent `\textit{passive updates}' of minority classes. Furthermore, our approach advances the performance of the ETF without any constraint on the metric space dimension, underscoring its capability to overcome the ETF constraint on the number of classes and metric space dimensionality.

In Table \ref{table5}, the advantage of the proposed method is further verified using ResNet-50 on ImageNet-LT. Concretely, our method outperforms PSC and ETF classifiers without having limitations on the metric space dimensionality and number of classes. These observations reflect the important role of dynamic label-to-prototype assignment in increasing metric space exploitation. 
Also, these observations show that our method is not limited to small-scale classification or backbone. Table \ref{table2} presents the results for evaluations on CIFAR-10, SVHN, and STL-10. Since the $d$ and $c$ are constant across these experiments, the same ${\bW}$ is used for these evaluations. The on-par performance of our method with the ETF classifier indicates that our approach to distributing prototypes on the hypersphere converges toward the ETF organization of prototypes, \ie, maximal separation, when the requirement of ETF is provided, \ie, $d>c$.

\begin{table}[]
\addtolength{\tabcolsep}{-1.0pt} 
\small
\begin{center}
\begin{tabular}{cccc}
\toprule
PSC  & ETF ($d=4096$) & Ours ($d=4096$) & Ours ($d=512$) \\ \midrule
44.3 & 44.7         & 44.6         & \textbf{44.9}     \\ \bottomrule  
\end{tabular}
\end{center}
\vspace{-16pt}
\caption{\small{ ImageNet-LT classification accuracy (\%) of the proposed and baseline prototypical methods using ResNet-50 as backbone.}} \label{table5}
\end{table}
\begin{table}[]
\addtolength{\tabcolsep}{-3.0pt} 
\small

\begin{center}
\begin{tabular}{lccc|ccc|ccc}
\toprule 

\multirow{2}{*}{} & \multicolumn{3}{c|}{CIFAR-10}                                                              & \multicolumn{3}{c|}{SVHN }                                                                  & \multicolumn{3}{c}{STL-10}                                                                \\ 
                        & 0.005                          & 0.01                           & 0.02                           & 0.005                          & 0.01                           & 0.02                           & 0.005                          & 0.01                           & 0.02                           \\ \midrule 
PSC                 & 67.3                           & 72.8                           & 78.6                           & 40.5                           & 40.9                           & 49.3                           & 33.1                           & \textbf{37.9} & \textbf{38.8} \\
ETF                     & \textbf{71.9} & 76.5                           & {81.0}    & \textbf{42.8} & 45.7                           & \textbf{49.8} & 33.5                           & 37.2                           & 37.9                           \\
Ours                    & 71.5                           & \textbf{76.9} & \textbf{81.4}                          & 40.9                           & \textbf{47.0} & 49.7                           & \textbf{35.7} & 35.6                           & 38.0                           \\ 
\bottomrule
\end{tabular}
\end{center}
\vspace{-16pt}
\caption{\small{ Long-tailed classification accuracy (\%) of the proposed and baseline prototypical methods using ResNet-32 when $d=64>c$ for imbalance factors of 0.02, 0.01, and 0.005. Note that the same ${\bW}$ is used for these evaluations. On-par results with ETF illustrate the convergence of our optimized prototype toward ETF structure when $d>c$.}} \label{table2}
\vspace{-10pt}
\end{table}

\subsection{Ablations}

\subsubsection{Hyperspherical Prototype Distribution}\label{HPD}
Prior to training, we distribute prototypes on $\small{{S}^{d-1}}$, and they remain fixed during the training. Hence, their separability critically affects the final discriminative power of the network. Following \cite{mettes2019hyperspherical}, we assess the distribution of prototypes by analyzing the maximum, and minimum pairwise cosine similarity among prototypes in Table \ref{table4}. This analysis is conducted with 100 prototypes when $\small{d=50}$ and $\small{d=100}$. 
In $\mathbb{R}^{50}$, our optimized prototypes achieve the best scores, reflecting the effectiveness of our approach in evenly distributing points on the hypersphere and providing large-margin separation among prototypes. Note that the dimensionality constraint of ETF requires that the metric space dimension should be at least equal to the number of prototypes. Therefore, the row for 100 ETF prototypes in $\mathbb{R}^{50}$ is empty.
In $\mathbb{R}^{100}$ our prototypes achieve the same scores as ETF, denoting the convergence of our method toward the ETF structure when $d\geq c$.

In Figure \ref{fig3}a, the advantage of our method in distributing hyperspherical prototypes is further verified
by illustrating the average pairwise cosine of 100 prototypes when $\small{d=c}$.
The distribution of prototypes in our method obtains a near-uniform distribution of inter-class cosine, similar to ETF, while HPN \cite{mettes2019hyperspherical} demonstrates a high level of variation which illustrates the efficacy of our method in uniformly distributing prototypes. It is worth noting that the minimal cosine similarity of the $\small{c}$ equiangular vectors in $\small{\mathbb{R}^d}$ is $\small{\frac{-1}{c-1}}$ \cite{papyan2020prevalence}, \ie, $\frac{-1}{100-1}\approx -0.01$ in this experiment.

\begin{table}[]
\small
\addtolength{\tabcolsep}{6pt} 
\begin{center}

 \resizebox{0.9\linewidth}{!}{
\begin{tabular}{lll|ll}
\toprule 
         & \multicolumn{2}{c|}{$\small{d=100}$}              & \multicolumn{2}{c}{$\small{d=50}$}               \\
         & Max          & Min                     & Max           & Min                  \\  \midrule
Word2Vec & 0.74        & -0.32                   & -             & -                       \\
HPN      & 0.05         & \textbf{-0.39}           &-0.05          &-0.62                  \\
ETF & \textbf{0.0} &-0.01&-&-\\
Ours     & \textbf{0.0} & -0.01          & \textbf{0.01} & \textbf{-1.0} \\ \bottomrule
\end{tabular}}
\end{center}
\vspace{-16pt}
\caption{\small{The maximum $\small{\downarrow}$, and minimum $\small{\downarrow}$ pairwise cosine similarity among 100 prototypes. 
The proposed distribution of prototypes obtains the best scores when $d<c$ and converges toward ETF geometrical structure when $d=c$.
}}\label{table4}
\vspace{-10pt}
\end{table}

\begin{figure}[]
\begin{center}
\includegraphics[width=0.9\linewidth]{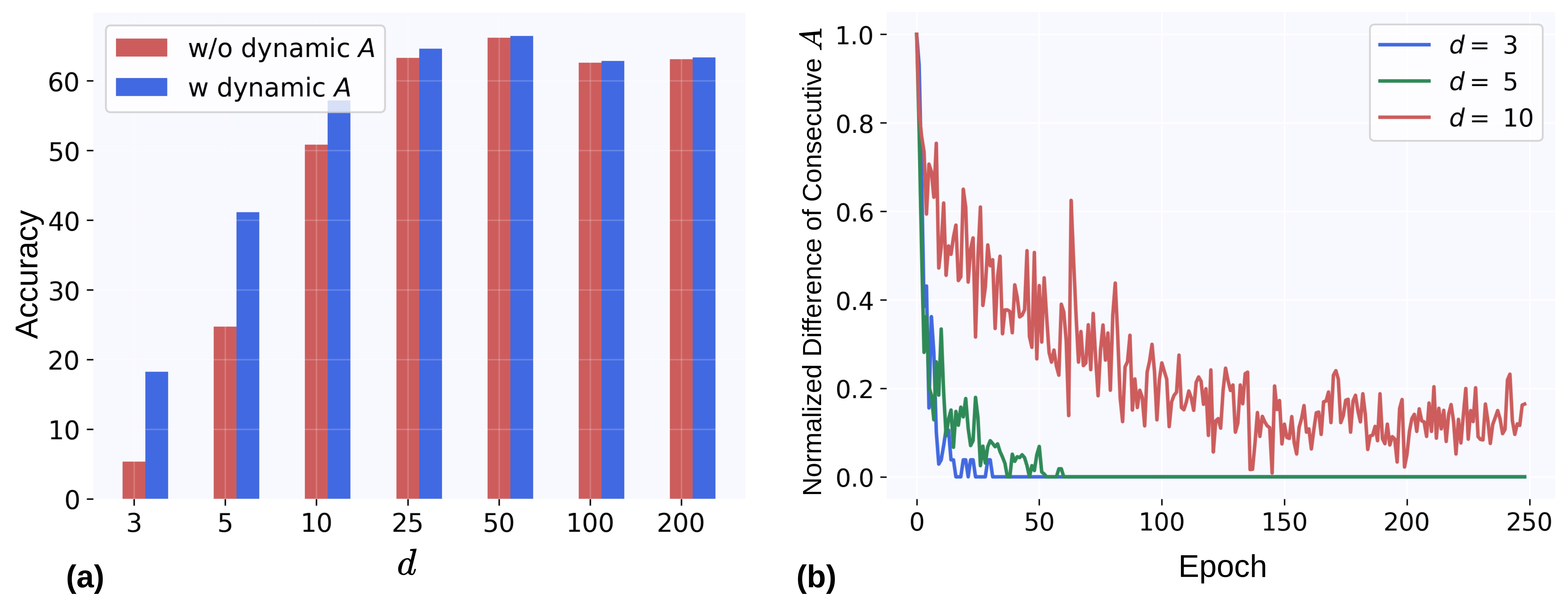}
\end{center}
\vspace{-16pt}
\caption{
a) Classification accuracy (\%) on CIFAR-100 using ResNet-32 w/wo dynamic assignment showing the significance of optimizing $A$ in low dimensional metric space.
b) Normalized difference of consecutive assignments during training.}\label{table6}
\vspace{-18pt}
\end{figure}

\vspace{-5pt}
\subsubsection{Ablation on $\tau$}\label{ablbm}
Here, we study the effect of the $\tau$ on the model performance.
For the sake of brevity, we define $\tau'=\frac{\tau}{\lfloor\frac{n}{\text{batch size}}\rfloor}$ which denotes the number of assignment updates per each epoch. 
Figure \ref{fig3}b shows the classification accuracy on CIFAR-100 when $\tau'$ changes from 0.1 to 10. We observe that $\tau'=10$ leads to performance improvement. However, with the increase in the $\small{d}$, the performance gap between $\tau'=10$ and $\tau'=1.0$ becomes reasonably close. 
Hence, to balance the computational efficacy and performance in the main experiments, we adhere to $\tau'=1$.

\subsubsection{Computational Time}
Computing label-to-prototype assignment imposes an additional workload that grows non-linearly with the increase in the number of classes, \ie., $O(c^3)$. 
Figure \ref{fig3}c reports the time consumed by a single label-to-prototype assignment computation. The reported value is the average of four runs. In our main experiments, we update the assignment mapping, $A$, at the end of every epoch, \ie, $\tau=\lfloor\frac{n}{\text{batch size}}\rfloor$. Hence, acquiring label-to-prototype assignment incurs negligible runtime overhead at the scale of 10,000 classes.

\vspace{-3mm}

\subsubsection{Impact of Prototype Distribution}
Previous studies \cite{duan2019uniformface,liu2018learning,wen2016discriminative} have employed the prototype uniformity constraint as the regularization to the Cross-Entropy (CE). In Table \ref{table15}, we compare our method with the regularized version of PSC, \ie, $\text{PSC*:}\text{ CE}+\lambda \: L_{uni}$. As expected, PSC* improves the PSC baseline performance. However, it introduces the challenge of finding optimal tuning parameter $\small{\lambda}$, which can vary across metric space dimensionality, depicted in Figure \ref{fig3}d. 
Additionally, we employed the best-performing PSC prototypes as the fixed $\bW$ of our method (referred to as Ours*). We observe that this model outperforms the baseline PSC model which suggests that our dynamic prototype matching provides notable enhancement to the training setup. Also, the intra-class compactness supervision of $\small{L_{IPM}}$ is important for this improvement. Note that our main model outperforms all these baselines due to the better exploitation of metric space.

\begin{table}[]
\addtolength{\tabcolsep}{7pt} 
\small
\begin{center}
 \resizebox{0.8\linewidth}{!}{
\begin{tabular}{lccc}
\toprule

                           & $\small{d=}$16                   & $\small{d=}$32                   & $\small{d=}$64                   \\ \midrule 
PSC     & 62.62   & 61.81 & 63.23 \\
PSC*                   &  62.54 & 64.18& 64.53 \\
Ours* & 60.51   & 63.43  & 64.71  \\
Ours              & \textbf{62.72}  & \textbf{65.09}   & \textbf{65.94}\\
\bottomrule                
\end{tabular}}
\end{center}
\vspace{-16pt}
\caption{\small{Abblation on the prototype distribution. The classification accuracy (\%) is reported on CIFAR-100 using ResNet-32. PSC*: $\text{CE} + 10^{-2}\times L_{uni}$. Ours*: substituting the equidistributed prototypes with softmax prototypes from a pre-trained network.}} \label{table15}
\vspace{-10pt}
\end{table} 

\vspace{-4mm}
\subsubsection{Impact of $A$}
Figure \ref{table6}a demonstrates the effect of label-to-prototype assignment in CIFAR-100 evaluation when $\small{d}$ varies from 3 to 200. When $d\ll c$, we observe that our optimizable assignment $A$ provides a notable enhancement to the training setup which emphasizes the importance of maintaining inter-class dependencies.
As $d$ increases, the effect of inter-class relationships fades. Also, we observe this effect in Figure \ref{table6}b. With the increase in $d$, the label-to-prototype mapping is converging to a constant mapping. These observations reflect that our method lifts the dimensionality constraint of ETF and converges toward ETF when the metric space requirements are provided. Thus, as the prototype distribution gradually converges toward ETF organization, the similarity between every pair of prototypes becomes equal, and inter-class dependencies become less important. 

\vspace{-3mm}

\section{Conclusion}
\label{sec:Conclusion}
In this paper, we proposed a dynamic label-to-prototype mapping to address the limitations of the current non-parametric alternative to the conventional Softmax classifier. Our method distributes prototypes before training and fixes them during training, to maintain the large-margin separation among prototypes and better metric space exploitation. 
Instead, we define the label-to-prototype assignment as an optimization variable, allowing the model to change the class that each prototype represents during training. 
The efficacy of the proposed method is demonstrated through various experiments on balanced/long-tail classification tasks and ablation studies.

\noindent\textbf{Acknowledgement.}
This research is based upon work supported by the Office of the Director of National Intelligence (ODNI), Intelligence Advanced Research Projects Activity (IARPA), via IARPA R\&D Contract No. 2022-21102100001. The views and conclusions contained herein are those of the authors and should not be interpreted as necessarily representing the official policies or endorsements, either expressed or implied, of the ODNI, IARPA, or the U.S. Government. The U.S. Government is authorized to reproduce and distribute reprints for Governmental purposes notwithstanding any copyright annotation thereon.

{
    \small
    \bibliographystyle{ieeenat_fullname}
    \bibliography{main}
}

\clearpage
\setcounter{page}{1}
\maketitlesupplementary

\section{Softmax Cross-Entropy Gradient Analysis}\label{fl}
With the identity mapping as the label-to-prototype assignment $A(y_i)=y_i$, the widely used softmax Cross-Entropy (CE) loss can be presented as follows:
\begin{equation}\label{softmax}
 \small
 \begin{aligned}
 L_{CE} = -\frac{1}{B}\sum_{i=1}^{B}{\log\frac{e^{\bz_i^{\top}\bw_{y_i}}}{e^{\bz_i^{\top}\bw_{y_i}}+\sum_{{j=1, j \neq y_i}}^c{e^{\bz_i^{\top}\bw_{j}}}}},
\end{aligned}
\end{equation}
where $\small{B}$ represents the batch size. The gradient with respect to the features can be expressed as:
\begin{equation}\label{featureGrad}
 \small
 \begin{aligned}
 \frac{\partial L_{CE}}{\partial \bz_i}\! =\! -(1\!\!-\!p_{y_i})\bw_{y_i}\!+\!\!\!\!\!\!\!\sum_{j=1,j\neq y_i}^c{\!\!\!\!\!\!p_j\bw_j}\!\!=\! {\bg}^{+} + {\bg}^{-},
\end{aligned}
\end{equation}
where $\small{p_j = \frac{e^{\bz_i^{\top}\bw_j}}{\sum_{k=1}^c{e^{\bz_i^{\top}\bw_k}}}}$.
Equation \ref{featureGrad} shows that from a feature perspective, the backbone optimization aims to align features with their corresponding ground-truth prototype, $\small{{\bg}^{+}}$, while ensuring separation from all other prototypes, $\small{{\bg}^{-}}$. Note that $\small{{\bg}^{-}}$ is necessary for a learnable classifier to impose the inter-class separability. However, in a fixed classifier framework, $\small{{\bg}^{-}}$ does not necessarily lead to separability and may cause performance degradation \cite{yang2022inducing}.

The gradient with respect to the prototypes can be expressed as:
\begin{equation}\label{classifierGrad}
 \small
 \begin{aligned}
 \frac{\partial L_{CE}}{\partial \bw_j} = -\sum_{i\in \mathbb{B}^+}{(1-p_i){\bz_i}}+ \sum_{i\in\mathbb{B}^-} {{p_i}{\bz_i}}= \hat{{\bg}}^{+}+\hat{{\bg}}^{-},
\end{aligned}
\end{equation}
$\mathbb{B}^+$ represents samples belonging to the $j$-th class in mini-batch, and $\mathbb{B}^-$ denotes the samples from other classes. Consequently, from a classifier perspective, the prototypes belonging to the $j$-th class are updated towards the features of samples from their own class, $\small{\hat{{\bg}}^{+}}$, while being pushed away from features of other classes, $\small{\hat{{\bg}}^{-}}$. 
Note that in Equation \ref{classifierGrad}, the $\small{\hat{{\bg}}^{-}}$ is available even when no sample from the class $\small{j}$ is present in the mini-batch (\textit{passive update}). Therefore, given an unbalanced dataset, the optimization of prototypes of minority classes is predominantly influenced by $\small{\hat{{\bg}}^{-}}$ \cite{deng2021variational}. Consequently, the prototypes of the minority classes gradually drift away from the feature space \cite{deng2021variational,yang2022inducing}. Additionally, $\small{\hat{{\bg}}^{-}}$ is approximately uniform across all minority classes, and forces prototypes of minority classes to the same subspace, leading to less separability among them \cite{deng2021variational}.

\begin{table}[]
\begin{center}
\addtolength{\tabcolsep}{-3pt} 
\small
\begin{center}
\begin{tabular}{lccc}
\toprule
\multicolumn{1}{c}{} & $\small{d=}$16         & $\small{d=}$32         & $\small{d=}$64         \\ \cline{2-4} \addlinespace
Fixed $\small{\bW}$ + CE   & 55.30  {$\pm$ 0.76} & 56.46 {$\pm$ 0.43} & 57.03 {$\pm$ 0.34} \\
Ours                 & \textbf{62.72}  {$\pm$ 0.36} & \textbf{65.09} {$\pm$ 0.24}   & \textbf{65.94} {$\pm$ 0.11}        \\  
\bottomrule
\end{tabular}
\end{center}
\end{center}
\vspace{-16pt}
\caption{\small{Abblation study on the effect of utilizing $\small{L_{IPM}}$ when the classifier prototypes are predefined and fixed during the training. The classification accuracy (\%) on CIFAR-100 using ResNet-32 is reported.}} \label{table14}
\vspace{-16pt}
\end{table} 

\section{Training Setups} \label{trainingsetup}

\subsection{Balanced}\label{hpnimpliment}
ImageNet-200 is a subset of ImageNet, containing 110K images from 200 classes. CIFAR-100 consists of 60K images from 100 classes. For both sets, 10K instances are randomly selected as the testing set and are held out from the training. 
We use SGD optimizer with batch size set as 128, initial learning rate as 0.01, momentum as 0.9, and weight decay as $\small{1e-4}$. All models are trained from scratch for 250 epochs, where in epochs 100 and 200, the learning rate is decreased by a factor of 10. For data augmentation, random cropping, and horizontal flips are conducted.

In ImageNet-1K experiments, ResNet-50 is trained using SGD optimizer with an initial learning rate of 0.5 with five epochs of linear warm-up and cosine decay learning rate. The batch size was set to 1024, weight decay to $2\times10^{-5}$ 0.00002, and total training epoch to 100. As the data augmentation, we employed Mixup with $\small{\beta}$ set as 1.0.
For Swin-T, we employed AdamW as the optimizer with a batch size of 1024, an initial learning rate of 0.001, and a weight decay of 0.00002.

\subsection{Long-tailed}
Models are trained from scratch for 200 epochs, using SGD optimizer with an initial learning rate of 0.1, a batch size of 128, momentum of 0.9, and weight decay of $2e-4$. The learning rate is decayed by a factor of 10 at epochs 160 and 180. 
Following ETF \cite{yang2022inducing}, the loss is weighted by the inverse ratio of the number of samples per class and in addition to random cropping and horizontal flipping, we used Mixup with hyperparameter $\small{\beta}$ set as 1.0.

\section{Additional Ablations}

\section{Impact of $L_{IPM}$}
In Table \ref{table14}, we study the effect of the $\small{L_{IPM}}$. To this end, we substitute the PSC prototypes with our optimized prototypes. In this experiment, the prototypes of PSC are fixed during the training. Employing CE as the training objective drastically degrades performance. 
This degeneration is due to the unnecessary pushing force in the gradient of CE as discussed in Section \ref{fl} of this Supplementary Material.
Also, this performance degradation emphasizes the important role of dynamic label-to-prototype assignment in better metric-space exploitation.

\end{document}